\documentclass{article}

\usepackage{arxiv, arxiv_te}

\makenomenclature

\title{DeepClean -- self-supervised artefact rejection for \unskip\space\ignorespaces intensive care waveform data using deep generative learning}

\author{\ignorespaces 
Tom Edinburgh\affils{1,2,5}\authorcontact{te269@cam.ac.uk}\orcid{https://orcid.org/0000-0002-3599-7133}\authsep 
Peter Smielewski\affils{3,5}\orcid{https://orcid.org/0000-0001-5096-3938}\authsep 
Marek Czosnyka\affils{3,5}\orcid{https://orcid.org/0000-0003-2446-8006}\\ 
Stephen J Eglen\affils{1,5}\orcid{https://orcid.org/0000-0001-8607-8025}\authsep 
Ari Ercole\affils{4,5}\orcid{https://orcid.org/0000-0001-8350-8093}}

\affil{\ignorespaces
\affils{1}Department of Applied Mathematics and Theoretical Physics\authsep
\affils{2}Cantab Capital Institute for Mathematics of Information\\
\affils{3}Brain Physics, Department of Clinical Neurosciences\authsep
\affils{4}Division of Anaesthesia, Department of Medicine\\
\affils{5}University of Cambridge, United Kingdom}

\begin{document}
\maketitle

\begin{abstract}
Waveform physiological data is important in the treatment of critically ill patients in the intensive care unit. Such recordings are susceptible to artefacts, which must be removed before the data can be re-used for alerting or reprocessed for other clinical or research purposes. Accurate removal of artefacts reduces bias and uncertainty in clinical assessment, as well as the false positive rate of intensive care unit alarms, and is therefore a key component in providing optimal clinical care. 
In this work, we present DeepClean; a prototype self-supervised artefact detection system using a convolutional variational autoencoder deep neural network that avoids costly and painstaking manual annotation, requiring only easily-obtained `good' data for training. For a test case with invasive arterial blood pressure, we demonstrate that our algorithm can detect the presence of an artefact within a 10-second sample of data with sensitivity and specificity around 90\%. Furthermore, DeepClean was able to identify regions of artefact within such samples with high accuracy and we show that it significantly outperforms a baseline principle component analysis approach in both signal reconstruction and artefact detection. DeepClean learns a generative model and therefore may also be used for imputation of missing data.
\end{abstract}


\suppressfloats

\section*{Introduction}

Critically ill patients in the intensive care unit (ICU) may experience life-threatening deterioration sometimes over minutes or even seconds. Care of these patients is therefore highly reliant on data \cite{Vincent2017-kl}, which, by necessity, may be voluminous and complex. An extreme example of this are high frequency physiological waveforms, such as invasive arterial blood pressure (ABP), the electrocardiogram and other vascular or intracranial pressures, which provide a wealth of complex, heterogeneous yet highly structured data at optimal sampling frequencies of 100Hz or even more. Continuous monitoring of these waveforms form the basis of clinical research and care, in particular alerting clinicians to changes in the patient state in real time, and they may be further processed to derive other useful parameters; for example, measures for tracking optimal cerebral autoregulation \cite{Aries2012-zb} and heart-rate variability \cite{Karmali2017-gq}, which has repeatedly been shown to be a determinant of physiological integrity. Further, more sophisticated metrics of these waveforms, based on non-linear dynamics \cite{Bishop2012-ri} or information theory \cite{Gao2017-bx},  may form novel digital biomarkers for precision care in the future (for example, \cite{Beqiri2019-iq}).

Waveform artefacts arise from a variety of internal and external sources, such as sensor noise, patient movement and clinical intervention. Examples of this include draining intracranial fluid and arterial flushing, when the arterial line is transduced to re-establish a relative pressure baseline. This is repeated at regular but infrequent intervals or upon moving the patient and decreases damping caused by blood clotting. Such artefacts may reduce reliability in estimation of these derived parameters, confound analysis and create uncertainty in clinical decision making. Closely related, in the same manner, is the handling of missing data, often treated using simple methods that are biased and underestimate variability, such as linear interpolation of observed data. Imputation methods that maintain some statistical properties or features of the data offer a useful alternative \cite{Sullivan2010-gh}. Artefact detection is also important for ICU alerting systems \cite{Scalzo2013-bf} and reduces the likelihood of false positive alarm incidents. A high rate of false alarms carries a significant risk as alarm fatigue often leaves clinical staff perceiving the alarms to be generally unhelpful and can therefore potentially result in delays to the appropriate clinical intervention \cite{Chambrin2001-gs}.

\begin{figure}[t]
  \centering
  \includegraphics[width=16cm]{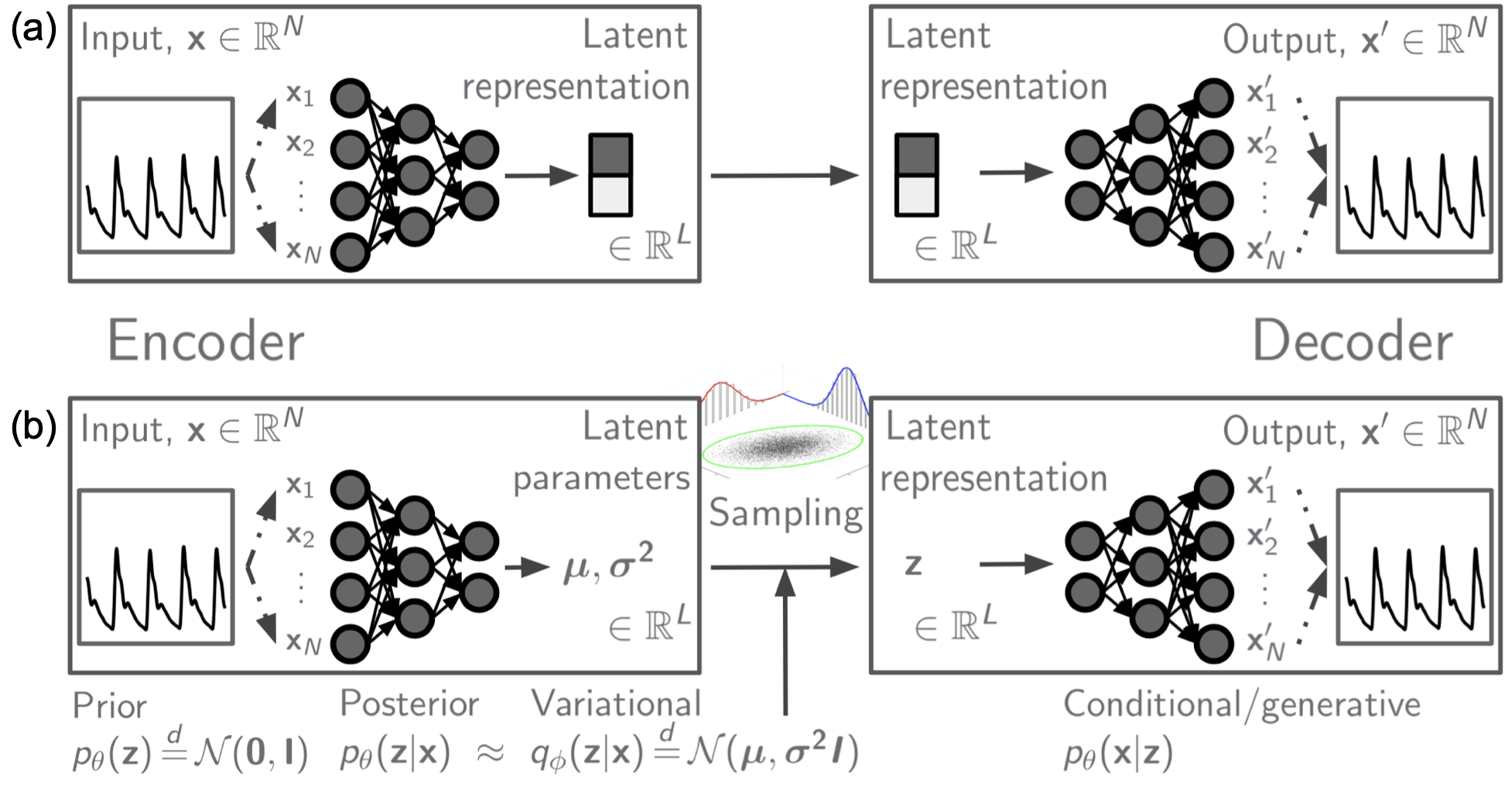}
  \captionsetup{justification=justified}
  \caption{(a) An autoencoder and (b) a variational autoencoder. An autoencoder network trains a parametric encoder and decoder simultaneously to learn a mapping that reconstructs the input via a sparse low-dimensional representation. The VAE instead learns a generative model, $p_\theta(x|z)$, and a variational approximation, $q_\phi(z|x)$, to the posterior distribution for the latent variables, $p_\theta(z|x)$, and we can generate new data by sampling from the latent and generative distributions in order. The weights of the encoder are variational parameters, $\phi$, are the weights of the decoder are generative parameters, $\theta$. The output of the encoder acts as parameters to the variational distribution, typically the mean and standard deviation of a Gaussian distribution.}
  \label{fig:autoencoders}
\end{figure}

Artefact detection has traditionally been a difficult and costly task, requiring time-consuming human annotation or thresholding based on signal-specific feature engineering \cite{Sun2006-sz}. Due to the complex morphologies of waveforms, annotation by experienced clinicians remains a gold standard for ICU multimodality monitoring, despite inherent biases and issues with replicability. Many standard supervised learning methods require samples that are annotated in this way in order to learn a model, though a recent study uses active learning to query and propose samples for annotation in an efficient manner \cite{Megjhani2019-ee}. An alternative unsupervised approach has foundations in spectral anomaly detection \cite{Chandola2009-bd}. These methods seek an embedding of the data in an `information bottleneck' lower-dimensional space where the anomaly or artefact is separated from the normal data, and then form a reconstruction of the input from its lower-dimensional representation back to the original input space. Embedding in the `bottleneck' latent space is a lossy transformation and the aim is to capture salient features of the data whilst disregarding anomalous features. Subsequently, the reconstruction should restore the underlying `true' behaviour and the error in the reconstruction, compared to the input, can then be used to discriminate artefacts. A classical autoencoder maps data inputs to a lower-dimensional sparse latent space and back to the original input space via encoder and decoder modules, which are both neural networks (Figure \ref{fig:autoencoders}a), so can be used for this purpose. However, its latent space is in general neither meaningful nor well-structured, and it may have issues with fragility and overconfident prediction \cite{Guo2017-uk}. Many extensions have been proposed to try to alleviate some of these problems \cite{Makhzani2013-eo, Vincent2008-un, Vincent2010-xo, Rifai2011-dw}, but classical autoencoders have largely fallen out of favour.

Generative modelling describes a class of models in which we want to learn a distribution $p_\theta(x)$ to approximate the true distribution of data $X$ in a dataset. Understanding the generative process behind some data is often very useful, as it helps us build useful abstractions and causal relationships are typically naturally embedded \cite{Kingma2019-an}. We can sample directly from this distribution to generate new example data that captures some statistical properties and features of the data. A subset of this are latent variable models, which condition on unobserved variables $z$, with a family of deterministic functions $f(z;\theta)$ mapping $z$ to the input data space via a decoder network, which is parameterised by $\theta$. The marginal $p(x)=\int p(x,z)dz=\int{p_\theta(x|z)p(z)dz}$ is also referred to as the model evidence; the probability of seeing the entire data under the generative model with parameters $\theta$. The aim is maximum likelihood estimation of $p(x)$ with respect to $\theta$, but this is often intractable. Variational autoencoders (VAEs) \cite{Kingma2013-om} resolve this problem using variational Bayesian inference together with probabilistic graphical models, reproducing this `bottleneck' architecture and borrowing the name `autoencoder' due to the presence of distinct but coupled encoder and decoder modules that give resemblance to the high-level structure of a classical autoencoder (Figure \ref{fig:autoencoders}b). The encoder describes a variational distribution $q_\phi(z|x)$ that approximates the true posterior $p(z|x)=p(x|z)p(z)/p(x)$, which itself would provide an ideal encoding of $x$ into $z$. During training, it learns both to encode some information about each data point through the latent representation $z\sim q_\phi(z|x)$ and stochastic sampling from this distribution allows the model to abstract and generate new candidate data $x\sim p_\theta(x|z)$ that accurately reconstructs inputs from the assumed underlying random process. This process enables it to discriminate artefacts despite never being explicitly exposed to any, since we assume different latent mechanisms govern the waveform behaviour of any artefacts, and so model prediction over an input containing such an artefact will have low reconstruction probability and high reconstruction error \cite{An2015-zo}. For further detail, comprehensive summaries of the VAE can be found in the literature e.g. \cite{Kingma2019-an}, \cite{Doersch2016-xa}, \cite{Stone2019-jd}.

In this work, we have demonstrated the potential of deep generative learning, in particular VAEs, in the detection of artefacts in physiological waveforms, developing a VAE-based model and pipeline, which we name DeepClean. It is worth emphasising that this method is unsupervised and thus doesn't require any previously annotated examples for training purposes, removing the need for costly human annotation at any stage of the process (apart from to assess performance of the model). We refer to it as self-supervised, as the deep generative learning alone doesn't itself provide binary labels marking artefacts but instead learns a reconstruction of the data, and therefore to classify artefacts, we need to augment this network with a decision rule or separate classification network. This final step involves the original input data via the reconstruction error, and may further include some self-evaluation to determine a decision boundary (for example, thresholding based on the input training data). We investigated whether, by training DeepClean on substantially clean ABP recordings as an example, regions of artefactual data could be identified in a self-supervised way without explicitly showing DeepClean any artefact exemplars. Furthermore, we aimed to see whether the DeepClean generative model could impute missing data over such regions.

\section*{Methods}

\subsection*{Data and preprocessing}

Fully anonymous arterial blood pressure (ABP) data from a standard indwelling arterial line connected to a pressure transducer (Baxter Healthcare Corp. CardioVascular Group, Irvine, CA) was obtained as part of routine ICU clinical care. The waveform was sampled at frequency of 125Hz using the ICM+ software (Cambridge Enterprise Ltd., Cambridge, UK, http://icmplus.neurosurg.cam.ac.uk). Under UK regulations, ethical approval was not required for the re-use of anonymous data obtained as part of routine clinical practice for research. 

\begin{figure}[t]
  \centering
  \includegraphics[width=16cm]{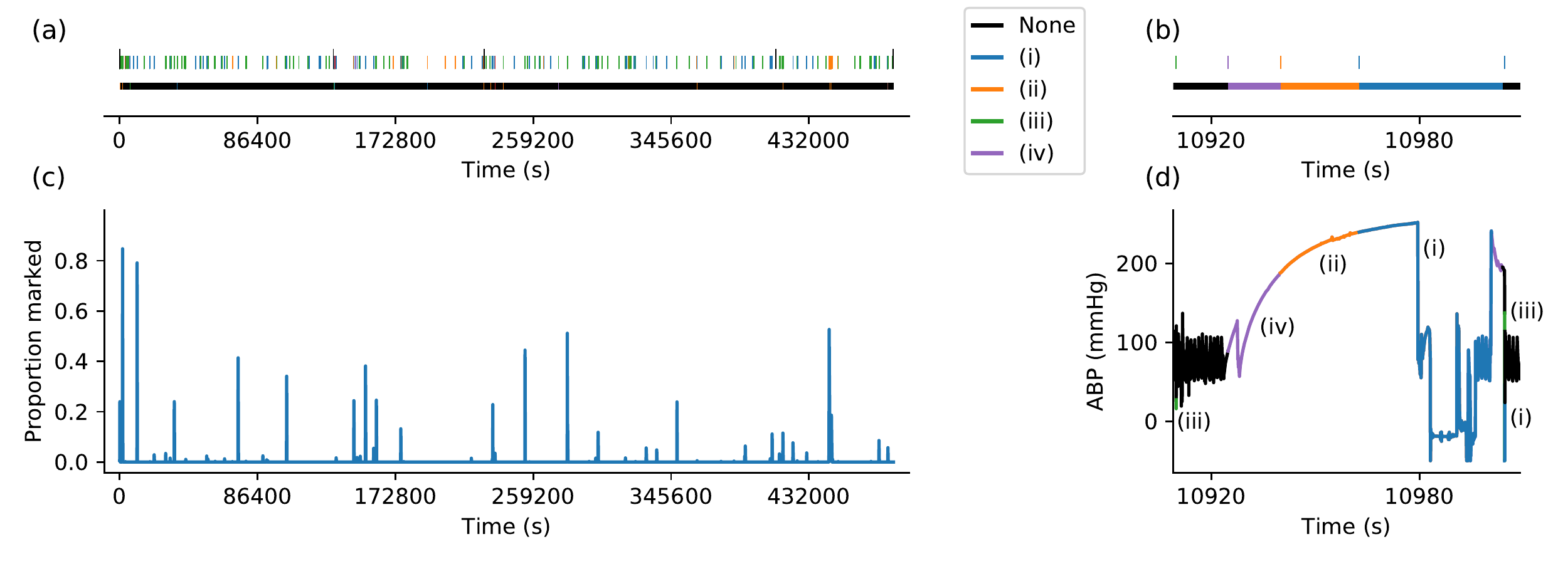}
  \captionsetup{justification=justified}
  \caption{Preprocessing: data was marked if (i) values exceeded sensible global, signal-specific thresholds, (ii) the range within a small sub-pulse time window stayed within a small threshold (i.e. the signal was static), (iii) an extreme local change in the signal within this sub-pulse window exceeded a much larger threshold, (iv) signal quality annotations, marked by clinical staff in real-time, were included within the dataset. In addition, successive marked regions were merged if the proportion of marked to not marked signal was particularly high (for example, (i) at 86,342s). (a) and (b) show the location of marked sections across the whole dataset and for an example subsection. As the widths of the marked sections are usually short relative to the dataset, a vertical line above also indicates the start of a marked section. There are several sections of missing data and longer vertical lines mark the start of each new recorded section. (c) shows the proportion of data within each 100-second window that is marked by this preprocessing and (d) illustrates marked data for the example subsection.}
  \label{fig:preprocessing}
\end{figure}

DeepClean was trained and tested on 10-second windows of preprocessed data. In order to learn a generative model describing `clean' physiological waveforms, we first removed large, grossly abnormal sections of the waveform by applying basic thresholding heuristics (Figure~\ref{fig:preprocessing}), to obtain substantially `clean' training data without human mark-up at a beat-to-beat level. This is a far easier task than manual markup of the waveform but inevitably fails to pick up all artefacts in the training set. However, for a typical waveform recording, the proportion of such abnormal data across the whole dataset will generally be very small (assuming signal artefacts are rare in comparison to normal beats). With sufficient quantities of data, it is possible to set conservative thresholds, such that the proportion of retained data containing an artefact is small, minimising their effect on the model. Removing this preprocessing step doesn't prevent the learning of a generative model but without it training can be more difficult and the final generative model sub-optimal, as training samples containing these abnormal sections are likely to have a larger contribution to the overall loss function. 

Though DeepClean is self-supervised, we required a labelled test set to determine model performance against manual expert mark-up. Further, we required this test set to be balanced, containing a similar proportion of artefacts and valid waveforms. We selected the test set randomly from the unprocessed data to avoid potential sources of selection bias, but with moderate bias towards regions marked as abnormal in preprocessing, as we assumed waveform artefacts were rare compared to normal beats across the entire data but regions marked as abnormal in preprocessing were likely to be enriched with artefacts. First, we split the data into 100-second windows and calculated the proportion of each window that had been marked in preprocessing (Figure \ref{fig:preprocessing}c). Normalising across the whole dataset to give a discrete probability measure, we sampled from this distribution to select a 100-second window that had greater chance of containing data marked as abnormal. We then sampled uniformly within this selected window to determine a 10-second sample to join the test set, repeating this procedure until we had generated a test set of sufficient size, in our case 200 10-second windows (or samples). The test set and preprocessing-marked regions were then removed from the dataset and the remaining data split into 10-second samples, shuffled, and divided into training and validation sets with ratio 9:1. All three data sets (training, validation and test) were then standardised by the training set mean and standard deviation. Each 10-second test sample was labelled as either containing an artefact or being artefact free by expert review (TE, AE) to enable us to assess the performance in artefact detection on a sample-wide basis. Furthermore, sections of the waveform \textit{within} each 10-second samples were also annotated to allow assessment of within-sample artefact detection performance. We also assessed performance by comparison to principal component analysis (PCA), which performs a similar dimensionality reduction and reconstruction to that previously described. PCA is related to a classical autoencoder in the simplest case with linear activations and the weights of the encoder network in this case span the same subspace as the principal components \cite{Cottrell1988-ij}. The number of PCA principal components is comparable to the latent dimension of a VAE, so refer to both as the latent dimension for brevity.

\subsection*{Model specification and training}

\begin{figure}[t]
  \centering
  \includegraphics[width=8.5cm]{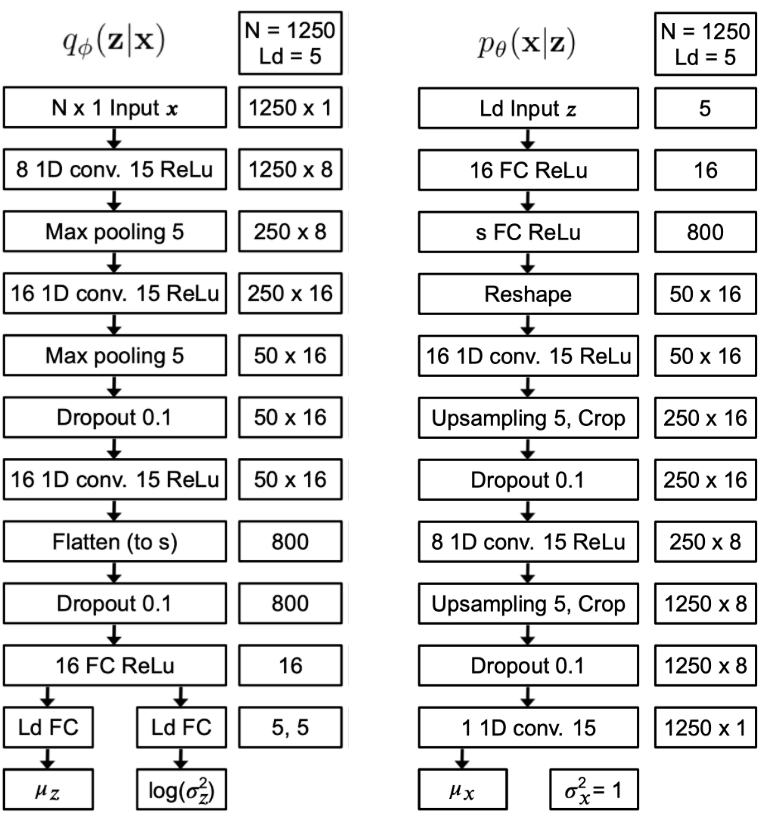}
  \captionsetup{justification=justified}
  \caption{DeepClean network architecture with tensor dimensions for example input and latent dimensions. This initial architecture was inspired by similar models in \cite{Kingma2013-om} and \cite{Chollet2017-iy}; see Discussion for further comments. Convolutional layers are described as `n 1D conv. m', where n is the number of convolutional filters and m the kernel size; pooling and upsampling layers show the pool size; dropout layers show the dropout rate; and dense fully connected (FC) layers show the number of units. Upsampling layers are accompanied by cropping of the end of the layer output to match the corresponding encoder dimension size. Activation functions, such as the `ReLu' activation, are shown where applicable; in all other cases, the activation is linear. `Ld' is the latent dimension.}
  \label{fig:network}
\end{figure}

We built a variational autoencoder with deep convolutional neural networks (CNNs) for both encoder and decoder modules. CNNs allow the network to learn translation-invariant local patterns, with successive layers building a spatial hierarchy of increasing scale, a sensible approach in this context as physiological signal data is quasi-periodic and highly-structured. We fixed an encoder network architecture with three relatively small convolutional layers, alternated with two pooling layers to increase the receptive field, and finally two dense layers that split the network graph into separate branches for the two parameters that define the latent distribution. The decoder architecture mirrored this in reverse, with pooling replaced by up-sampling (Figure~\ref{fig:network}). Both encoder and decoder contained approximately 20,000 trainable parameters. We then trained separate models with increasing latent space dimension. For each latent dimension, we repeated training five times, presenting the model that minimised the validation loss. To train DeepClean, we used an NVidia Pascal P100 GPU, using Python with the Keras library \cite{Chollet2015-xk} and TensorFlow backend \cite{Abadi2016-ua}.

We optimised the evidence lower bound (ELBO) objective \cite{Kingma2019-an, Kingma2013-om}. Concurrently optimising over $\theta$ and $\phi$, the ELBO achieves important dual goals: maximising the marginal distribution $p(x)$ and minimising the divergence between the variational distribution $q_\phi(z|x)$ and the true unknown posterior $p(z|x)$. We made standard assumptions about the distributional families of the prior, variational and conditional. For instance, we assumed the prior distribution was a standard multivariate Gaussian, $p(z) =^d N(0, I)$, and the variational distribution was a factorised multivariate Gaussian with diagonal covariance matrix, $q_\phi(z|x) =^d N(\mu(x), \sigma^2(x) I)$, where $\mu(x)$ and $\sigma^2(x)$ are deterministic functions of $x$ computed through the encoder network. In theory, this diagonal covariance encourages independent features and hence an efficient encoding, though the picture is much more complicated in practice and the network can be susceptible to `pruning' some latent dimensions which then encode no information about the data \cite{Sonderby2016-dt}. Given our dataset was of continuous real-valued inputs, we also modelled the generative conditional distribution as a high-dimensional multivariate Gaussian distribution $p_\theta(x|z) =^d N(\mu(z), \Sigma(z))$, parameterised by the decoder network weights $\phi$. Often, the covariance of this distribution is taken to be $I$, for reasons discussed in \cite{Shu2018-tj}, especially since we are generally only interested in the mean $\mu(z)$. In the results we present here, we made this final assumption about the covariance $\Sigma(z)$, though we also separately investigated relaxing this assumption and allowing more general $\Sigma(z)$. The assumptions we made typically do not reduce the learning capacity of a VAE model, provided the model is sufficiently non-linear, and, together with Monte Carlo stochastic approximations to the expectation terms, provide a tractable expression for the gradient of the objective and result in a well-structured latent distribution (the average encoding distribution \cite{Hoffman2016-bw}, Figure \ref{fig:latent_examples}). Stochasticity helps regularisation of the latent space and allows more robust and meaningful latent representations, since it forces the variational model to place higher probability on a range of $z$ values that could potentially have generated the input sample, which prevents a collapse down to a single point estimate, and therefore similar inputs have similar representations and vice versa. This is illustrated by Figure~\ref{fig:latent_examples}, which highlights the robustness of the generative model in that minor perturbations to the input will not drastically change either the latent representation or the final output (Figure~\ref{fig:latent_examples}a, top row).

\subsection*{Sample-wide and within-sample artefact detection}

We assessed the performance of our approach in detecting whether a 10-second sample of data contained an artefact or not using the mean squared error (MSE) between the sample and its reconstruction, with a suitable threshold. The goal of this work was to develop a `blind' self-supervised classification procedure, and so any threshold on a metric applied to the reconstruction error must therefore be chosen independently of the test data. Figure \ref{fig:logmse} illustrates that values of metrics are generally not independent of hyperparameter choice. Therefore, without prior knowledge of a suitable threshold, a pragmatic approach was to set a threshold based on the 90\textsuperscript{th} percentile of the same metric calculated instead on the training data and their corresponding reconstructions. By decreasing the threshold on a metric, we classify more samples as artefacts, regardless of the correctness of this classification, and will therefore increase the specificity at the cost of decreasing the sensitivity (Figure \ref{fig:logmse}). We assessed the model performance by comparing the DeepClean classification to our annotation, using measures widely employed in clinical settings: accuracy, sensitivity and specificity.

With samples identified as artefacts, we can then use a metric on subsets of the sample (such as a sliding window approach) to identify artefacts more precisely within these samples; we also calculated the MSE on a 1-second sliding window (physiologically similar to one or two typical heartbeat periods) for this purpose. We made a distinction that if the MSE on a particular window exceeded the threshold, then the entire window was identified as an artefact, and not just its midpoint. We defined a threshold for identifying an artefact as before, but used the training set 99\textsuperscript{th} percentile here. We provide justification of using this more stringent threshold in the Discussion. We assessed the performance by considering the proportion of each sample correctly identified with respect to our annotation. In addition, we considered the proportion of artefact within each sample that is correctly assigned as artefact, and similarly the proportion of the non-artefact sections that DeepClean does not incorrectly assign as artefact. In these last two measures, we ignored samples that contain no artefact and samples that contain only artefact respectively. These are analogous to the binary classification measures.

\section*{Results}

We work with ABP waveform data from a single anonymised (adult) patient monitored almost continuously throughout a stay of several days (486,984 seconds) in the ICU. We marked 11,082 seconds of 486,984 seconds (2.28\%) as grossly abnormal in preprocessing, leaving training and validation sets of 37,821 and 4,728 10-second samples. Training required under 10 minutes of computation time on average, although this increased with latent dimension. Subsequent prediction, on test data or on new data, is inexpensive and requires only milliseconds. Our annotation marked 130 test samples out of a test set of 200 as containing an artefact.

\begin{figure}[t]
  \centering
  \includegraphics[width=16cm]{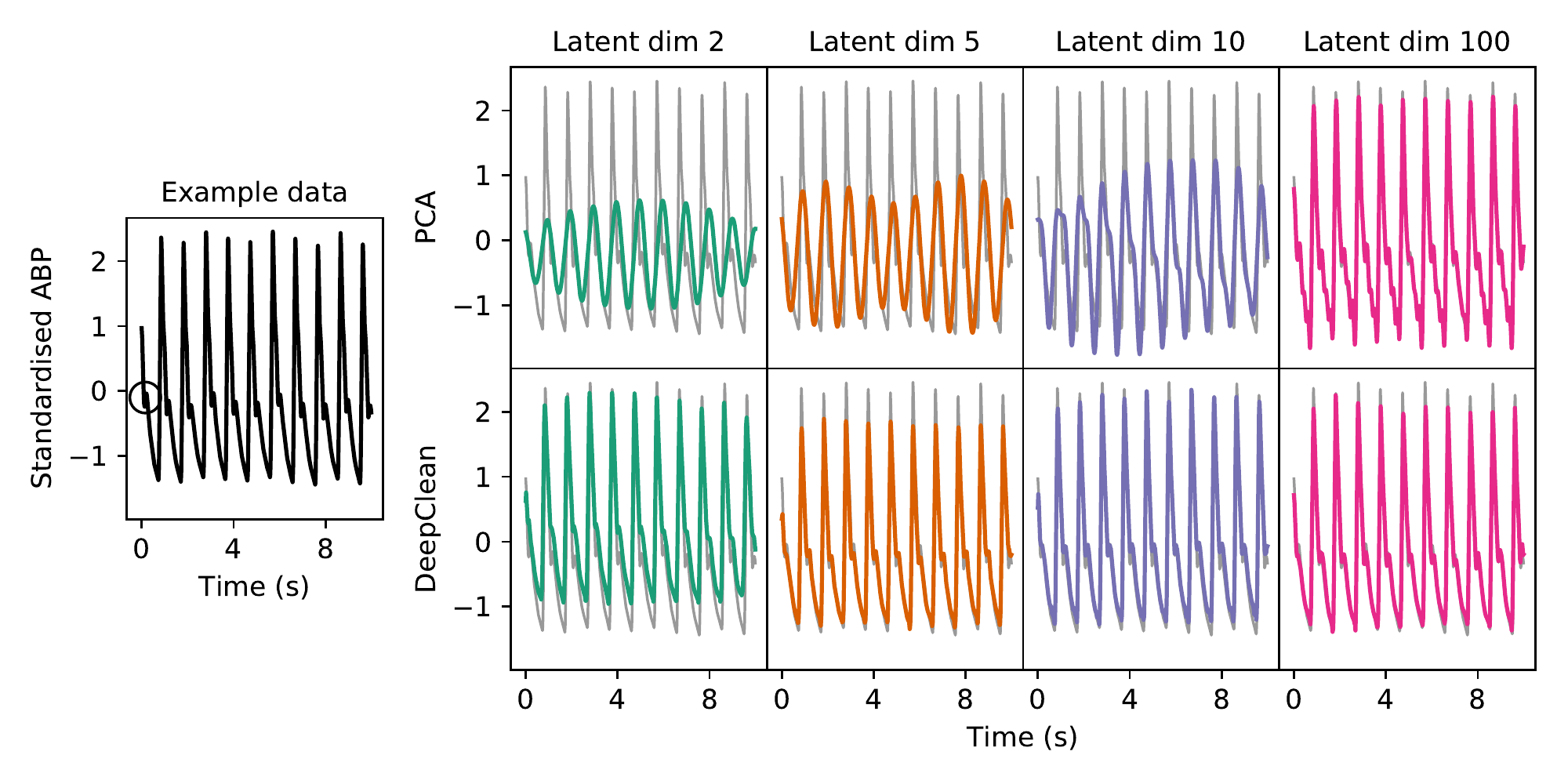}
  \captionsetup{justification=justified}
  \caption{Example 10-second sample data, with PCA (top) and DeepClean (bottom) reconstructions for increasing latent dimension (or PCA number of principal components). The original signal is shown in grey and the reconstruction in colour. PCA performs poorly unless the number of principal components is large, whereas DeepClean can encode sub-pulse components even with minimal latent information. The dicrotic notch and diastolic peak of an ABP beat are circled.}
  \label{fig:reconstruction}
\end{figure}

\begin{figure}[t]
  \centering
  \includegraphics[width=16cm]{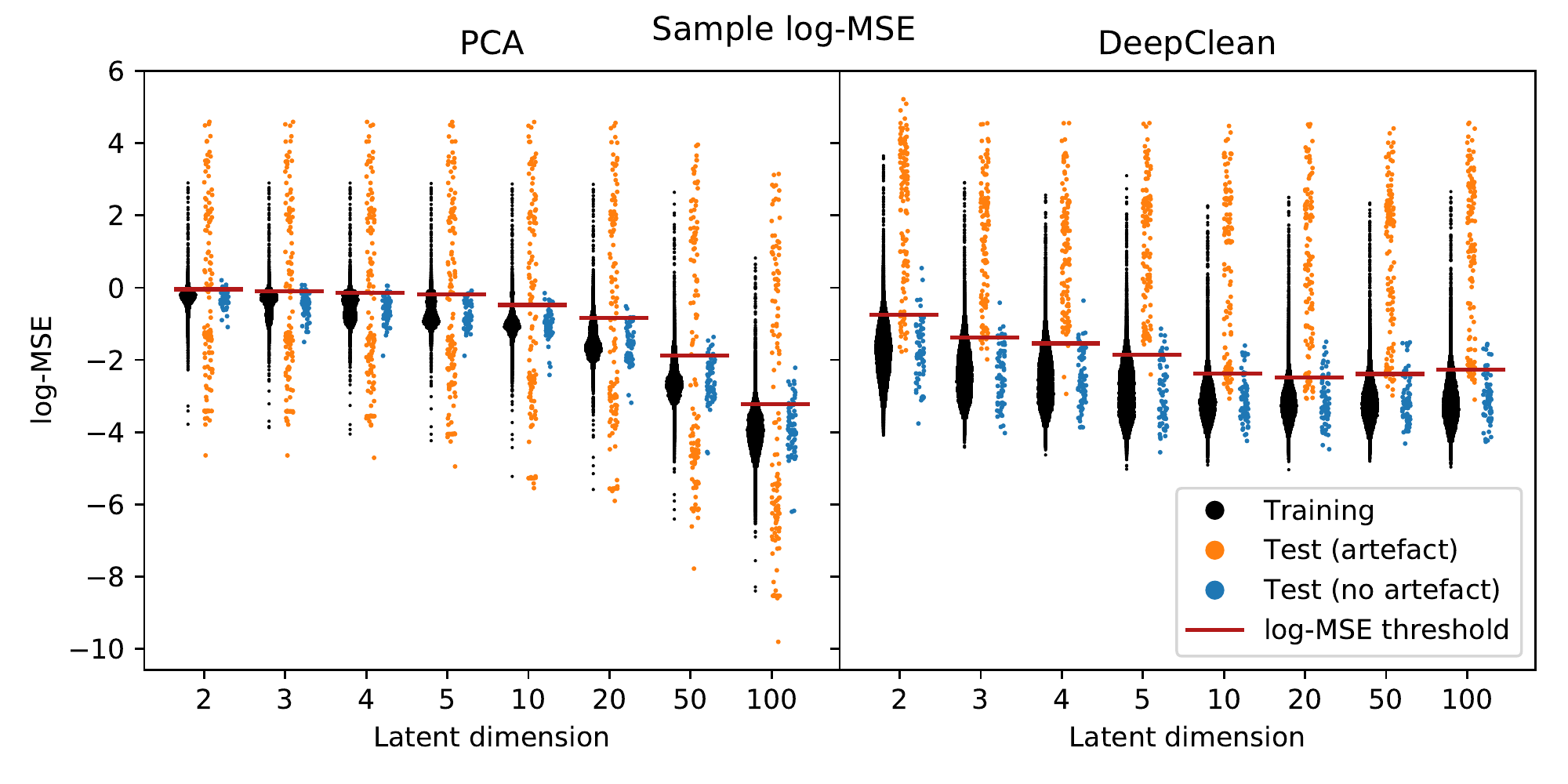}
  \captionsetup{justification=justified}
  \caption{Log-MSE values for both PCA and DeepClean, with increasing latent dimension. The distribution of training set log-MSE is shown with the training set 90\textsuperscript{th} percentile threshold. The test samples are split into those annotated as artefacts and not, and shown separately. The `ground truth' artefacts that lie below the threshold and non-artefacts above the threshold were samples that were incorrectly identified, and were false negatives and false positives respectively. Therefore, the proportion of test artefacts points above the threshold is the sensitivity and the proportion of test non-artefacts below the threshold is the specificity.}
  \label{fig:logmse}
\end{figure}

\begin{figure}[t]
  \centering
  \includegraphics[width=16cm]{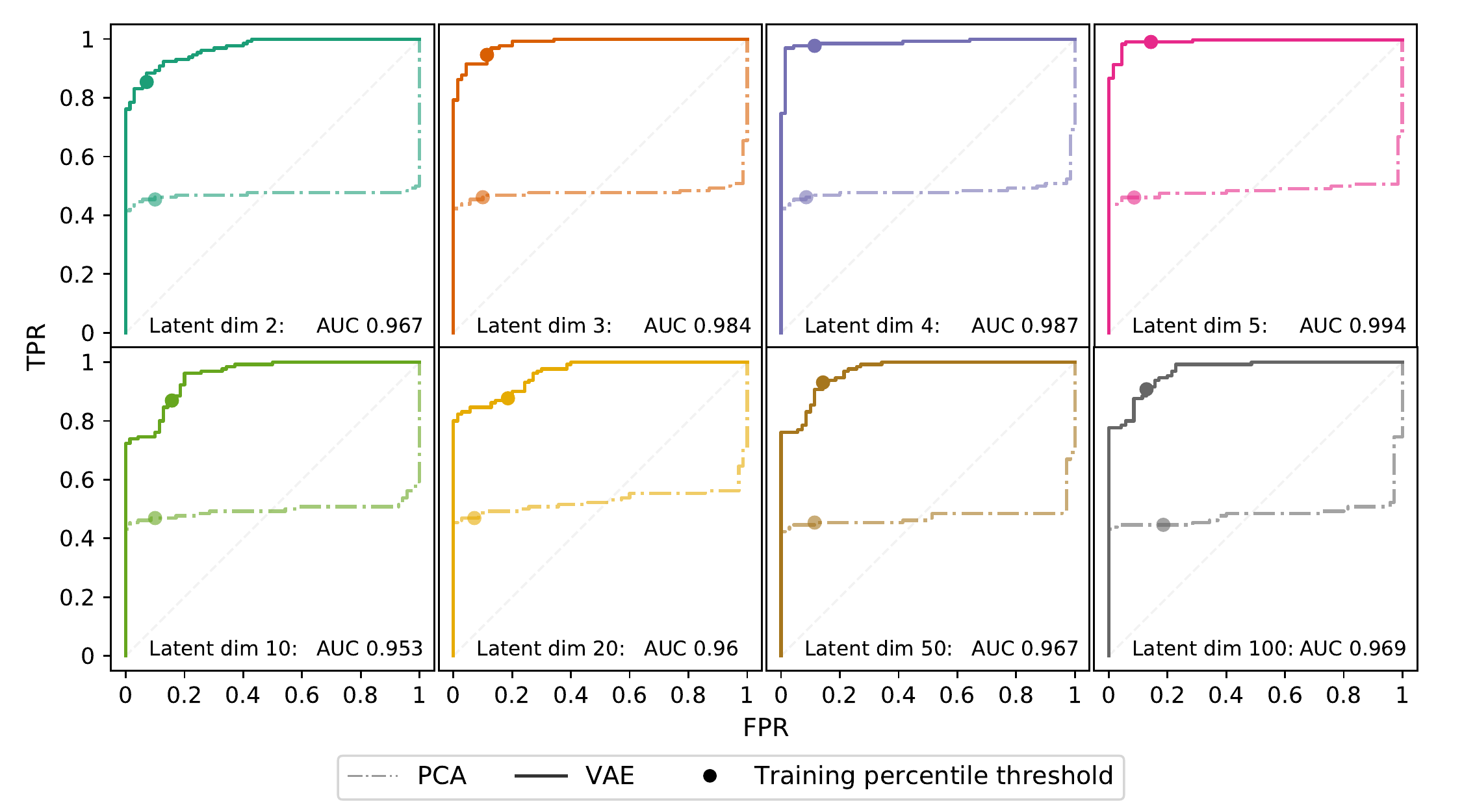}
  \captionsetup{justification=justified}
  \caption{Receiver operating characteristic (ROC) curves for each latent dimension. True positive rate (TPR) and false positive rate (FPR) are the same as sensitivity and (1 - specificity) respectively. The points marked on each curve correspond to our training set percentile threshold. The area under the curve (AUC) for DeepClean is also given.}
  \label{fig:roc}
\end{figure}

DeepClean was able to much more accurately reconstruct the waveform inputs than PCA (Figure \ref{fig:reconstruction}). In particular, DeepClean was able to encode sub-pulse components, such as the dicrotic notch and subsequent diastolic peak, even for minimal latent dimension. The PCA reconstruction is particularly poor for a small number of principal components but improves for much larger latent dimensions (as seen in the training set log-MSE values in Figure~\ref{fig:logmse}). However, for a high number of principal components, the PCA reconstruction contains high-frequency elements not present in the signal or the smoother DeepClean reconstruction. DeepClean substantially outperformed the baseline PCA in sample-wide artefact detection in model sensitivity (Table \ref{tab:samplewide}), using our training-set defined threshold. Whilst we give this heuristic for identifying a suitable threshold, we also provide receiver operating characteristic (ROC) curves to show the effect of varying the threshold (Figure \ref{fig:roc}). DeepClean had a significantly higher ROC area under the curve (AUC) than PCA in all cases. There is in general a clear distinction between the log-MSE of artefacts and valid data for the DeepClean reconstruction, since DeepClean is able to distinguish samples similar to and unlike those belonging to the underlying generative process of the waveform. In contrast, the range of log-MSE values for PCA reconstructions were similar for both training set samples and test set artefacts, so the latter cannot be easily identified in this way. One explanation for this is that regularisation term of the VAE enforces a localised latent distribution and penalises new samples sufficiently different enough from those in the training set that they have representations outside of this distribution. This does not happen in PCA and so there is no mechanism in PCA that prevents such samples from having a sufficiently accurate reconstruction, despite them being unlike previously seen data. In both cases, the non-artefact test samples followed a similar distribution to the training data, so the specificity was generally close to the training set 90\textsuperscript{th} percentile threshold. As PCA reconstructions have higher error than DeepClean unless the latent dimension is very large, the threshold was different for each method and as a result identical reconstructions may be classified differently by DeepClean and PCA. In particular, a DeepClean reconstruction of a sample containing an artefact that has high error may be identified as such when an identical reconstruction from PCA may not.

\begin{table}[t]
  \centering
  \begin{tabular}{ccccccccc}
    \toprule
     &  \multicolumn{2}{c}{Accuracy} & \multicolumn{2}{c}{Sensitivity} & \multicolumn{2}{c}{Specificity} & \multicolumn{2}{c}{ROC AUC} \\
    \cmidrule(r){2-3}
    \cmidrule(r){4-5}
    \cmidrule(r){6-7}
    \cmidrule(r){8-9}
    Latent dim & PCA & VAE & PCA & VAE & PCA & VAE & PCA & VAE \\
    \midrule
    \hphantom{00}2 & 0.61\hphantom{0} & 0.88\hphantom{0}  & 0.454 & 0.854 & 0.9\hphantom{00} & 0.929 & 0.471 & 0.967 \\
    \hphantom{00}3 & 0.615 & 0.925 & 0.462 & 0.977 & 0.9\hphantom{00} & 0.886 & 0.478 & 0.984 \\
    \hphantom{00}4 & 0.62\hphantom{0} & 0.945 & 0.462 & 0.977 & 0.914 & 0.886 & 0.482 & 0.987 \\
    \hphantom{00}5 & 0.62\hphantom{0} & 0.945 & 0.462 & 0.992 & 0.914 & 0.857 & 0.487 & 0.994 \\
    \hphantom{0}10 & 0.62\hphantom{0} & 0.86\hphantom{0} & 0.470 & 0.869 & 0.9\hphantom{00} & 0.843 & 0.496 & 0.953 \\
    \hphantom{0}20 & 0.63\hphantom{0} & 0.855 & 0.470 & 0.877 & 0.929 & 0.814 & 0.528 & 0.96\hphantom{0} \\
    \hphantom{0}50 & 0.605 & 0.905 & 0.454 & 0.931 & 0.886 & 0.857 & 0.474 & 0.976 \\
    100 & 0.58\hphantom{0} & 0.895 & 0.446 & 0.908 & 0.828 & 0.871 & 0.478 & 0.969 \\
    \midrule
    Mean & 0.613 & 0.901 & 0.460 & 0.919 & 0.896 & 0.868 & 0.487 & 0.973 \\
    \bottomrule \\
  \end{tabular}
  \captionsetup{justification=justified}
  \caption{Assessment of the classification of samples as containing an artefact or not, for both PCA and DeepClean (VAE). ROC AUC is the area under the receiver operating characteristic curve (Figure \ref{fig:roc}). Both methods had comparable specificity, i.e. correctly identifying non-artefact data as such, with PCA performing slightly better. However, DeepClean alone can identify artefactual data as such, with high sensitivity.}
  \label{tab:samplewide}
\end{table}

The second task of a more precise identification of artefacts within each sample is a more difficult problem. Both PCA and DeepClean performed less well in this (Table \ref{tab:withinsample}). For a significant number of samples in the test set, the entire sample was clearly either entirely artefact or entirely valid waveform. For almost half of the test samples, the entire sample was classified completely (100\%) correctly by DeepClean across all latent dimensions, corresponding to most of these cases. PCA was able to classify some of these samples completely correctly but also completely incorrectly classified others. Some examples of the DeepClean artefact detection compared to our annotation are provided in Figure \ref{fig:latent_examples}. We discuss later the difficulties in this task and relate this to the latent representations of these samples.

\begin{table}[t]
  \centering
  \begin{tabular}{ccccccccc}
    \toprule
    & \multicolumn{6}{l}{Mean proportion correctly identified of the:} & & \\
    \cmidrule(r){2-7}
    & \multicolumn{2}{c}{Entire sample\hphantom{0000}} & \multicolumn{2}{c}{Artefact w.s.\hphantom{0000}} & \multicolumn{2}{c}{Non-artefact w.s.} & \multicolumn{2}{c}{Prop. 100\% correct} \\
    \cmidrule(r){2-3}
    \cmidrule(r){4-5}
    \cmidrule(r){6-7}
    \cmidrule(r){8-9}
    Latent dim & PCA & VAE & PCA & VAE & PCA & VAE & PCA & VAE \\
    \midrule
    \hphantom{00}2 & 0.507 & 0.818 & 0.779 & 0.896 & 0.950 & 0.943 & 0.235 & 0.595 \\
    \hphantom{00}3 & 0.500 & 0.818 & 0.778 & 0.889 & 0.945 & 0.951 & 0.22\hphantom{0} & 0.555 \\
    \hphantom{00}4 & 0.504 & 0.796 & 0.779 & 0.893 & 0.956 & 0.949 & 0.235 & 0.49\hphantom{0} \\
    \hphantom{00}5 & 0.505 & 0.820 & 0.776 & 0.889 & 0.944 & 0.949 & 0.26\hphantom{0} & 0.56\hphantom{0} \\
    \hphantom{0}10 & 0.530 & 0.752 & 0.796 & 0.857 & 0.946 & 0.94 & 0.315 & 0.51\hphantom{0} \\
    \hphantom{0}20 & 0.551 & 0.794 & 0.807 & 0.853 & 0.946 & 0.965 & 0.35\hphantom{0} & 0.51\hphantom{0} \\
    \hphantom{0}50 & 0.545 & 0.777 & 0.772 & 0.853 & 0.949 & 0.955 & 0.34\hphantom{0} & 0.57\hphantom{0} \\
    100 & 0.556 & 0.776 & 0.926 & 0.755 & 0.982 & 0.965 & 0.375 & 0.54\hphantom{0} \\
    \midrule
    Mean & 0.525 & 0.794 & 0.780 & 0.873 & 0.951 & 0.956 & 0.291 & 0.541 \\
    \bottomrule \\
  \end{tabular}
  \captionsetup{justification=justified}
  \caption{Assessment of the classification of artefacts within each sample, for both PCA and DeepClean (VAE). For each sample, we calculated the proportion of that sample that was correctly identified with respect to our manual annotation, and we report the average across all samples. In addition, we show the proportion of artefactual data within the sample that is correctly identified as being an artefact, and similarly for non-artefactual data (with `within sample' abbreviated to `w.s.'). Finally, we show the proportion of samples within which the entirety of the sample was correctly (100\%) identified.}.
  \label{tab:withinsample}
\end{table}

\section*{Discussion}
\label{sec:discussion}

We have demonstrated the ability of a VAE to clean ABP signals in a self-supervised manner, suggesting a clear potential role for deep generative learning within this clinical research field and application. In addition to jointly training the encoder and decoder neural networks, DeepClean requires only two basic steps: straightforward thresholding heuristics for preprocessing, which may even be omitted at the cost of a slightly weaker model, and a metric or decision rule for discriminating artefacts based on the reconstruction error, which may include learning an acceptance threshold based on an approximate target false negative rate. Unlike a recent study \cite{Lee2019-hr}, which employed a stacked convolutional autoencoder (SCAE) combined with a CNN, our approach does not require pulse pre-segmentation (which is in itself a difficult task requiring heuristics) or a supervised final classification network. Additionally, since DeepClean functions in the time domain, it avoids the quadratic complexity of first mapping the pulse morphology to a 2-dimensional space.

We decided to split the data into 10-second samples for training and prediction, since such a segment will typically contain a small number of beats and we do not expect physiology to vary grossly over this period from clinical experience. A sample of this length should include a sufficient number of beats that the model can learn and generalise the structure of the waveform, and therefore distinguish artefacts that occur over similar or longer timescales. It is important that the model recognises clinical events as part of the same generative process, and so does not classify these as artefacts. Failure to flag a clinical event that has been incorrectly identified as a waveform artefact risks delayed intervention and treatment. Whilst such events are characterised by sudden changes across multiple frequency bands concurrently, the structure of the waveform during a clinical event is largely retained within a sample of this length, so it is a suitable choice.

\begin{figure}
  \centering
  \includegraphics[width=17.5cm]{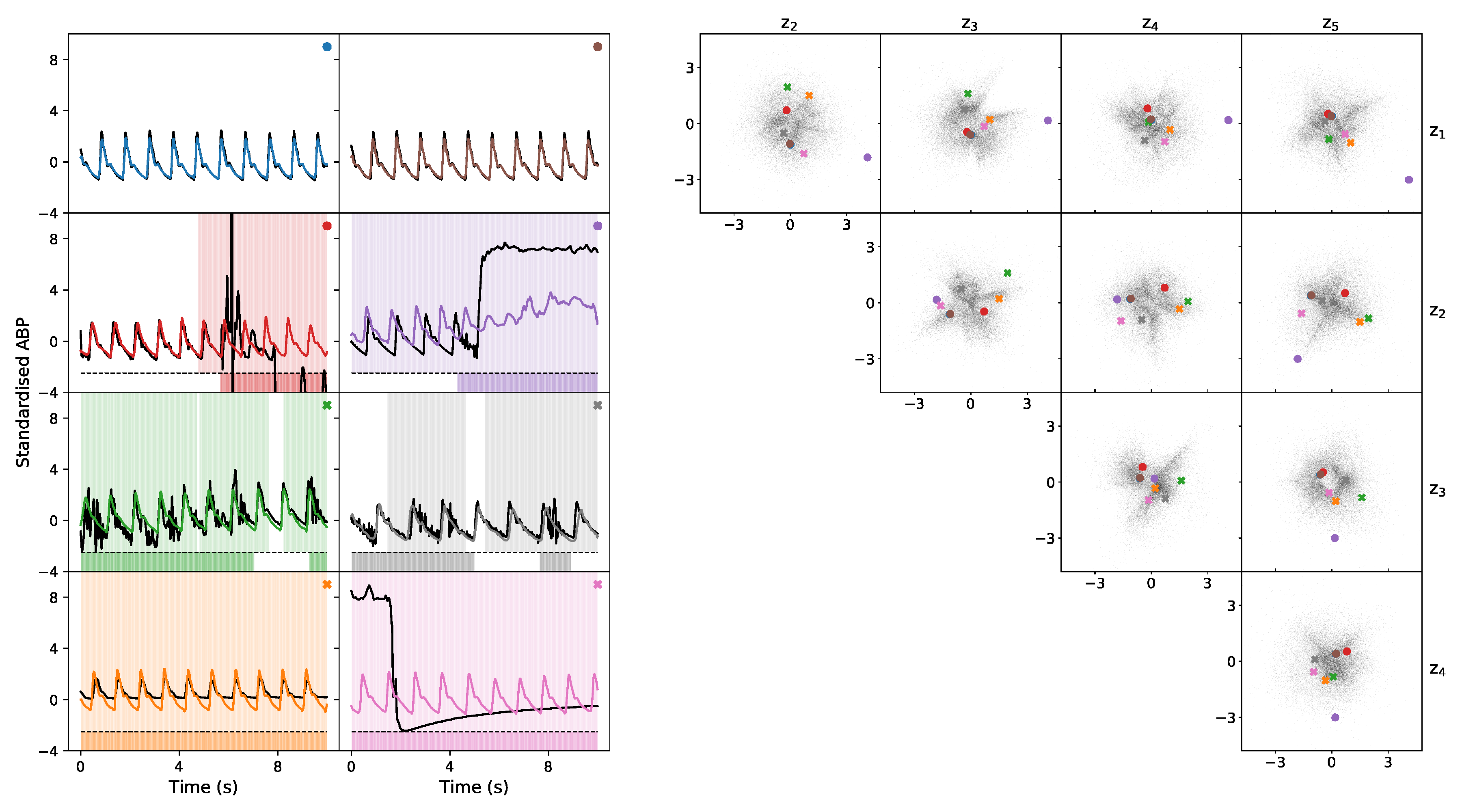}
  \captionsetup{justification=justified}
  \caption{Example input, representations and reconstructions, from the same DeepClean model with latent dimension 5, illustrating key principles involving the latent distribution. (a) The input and reconstructions are shown on the left and (b) the latent representations on the right; each subplot shows the distribution of a pair of latent variables, after marginalising over all other latent variables. This latent distribution, which is called the average encoding distribution, is approximately equal to the distribution of the prior \cite{Hoffman2016-bw}, a multivariate unit Gaussian. Each example test input is shown in black (a) with the corresponding latent representation within the latent distribution (b) and reconstruction (a) in (matching) colour. The top two examples are from the training set and the rest from the test set, with the within-sample artefact detection shown. The shaded region above the dotted line (a) shows where DeepClean identifies an artefact and the shaded region below the dotted line shows the artefactual region as defined by the manual annotation. Two different marker types are used for the representation for ease of reading and the marker type for each representation is shown in the corner of the input plot. These example include noisy data (third row), attenuation (left column, fourth row) and attenuation (right column, second and fourth row).}
  \label{fig:latent_examples}
\end{figure}

We sought a principled approach in selecting a suitable cutoff for any metric defined on the reconstruction error and used to determine an artefact, given that this choice appears dependent on hyperparameters such as the latent dimension. By decreasing the threshold on a metric, we classify more samples as artefacts, regardless of the correctness of this classification, and will therefore increase the specificity at the cost of decreasing the sensitivity (Figures \ref{fig:logmse} and \ref{fig:roc}). We set a more stringent threshold for the task of identifying artefacts within each sample. We reasoned that it is useful practically to more closely restrict the number of false positive identifications in this setting, since, for example, false identification of 10\% of each sample results in significantly more instances of false artefacts than false identification of 10\% of samples, and is therefore more difficult to handle or check. In addition, if we consider the distribution of MSE training set values in both cases, since the MSE averages the reconstruction error over a smaller number of time points for a 1-second window than for a 10-second window, any artefact is in general likely to lie further towards the tail of this distribution in the former compared to the latter, as the contribution towards the MSE of a single artefactual value is higher when the number of data points is smaller. For example, the left column, second row test set sample in Figure \ref{fig:latent_examples}a was classified as containing an artefact in our first test, even though the first half of the sample contributes little to the overall MSE across the entire sample. In contrast, 1-second window MSE values in the second half of the sample are much larger as there is no contribution from the valid non-artefactual first half of the sample; therefore we can afford a stricter threshold to determine these artefactual regions.

Artefact detection is often only the first of two parts involved in handling invalid data, as simply removing artefacts creates missing data that may also bias further analysis. One major advantage of a deep generative model such as a VAE is that we can generate realistic, synthetic data post-training, by sampling directly from the latent distribution. An obvious solution then is to replace samples containing an artefact with their reconstruction, and further, we can also set missing data to a fixed non-viable value and treat this as an artefact similarly. This may only be suitable for a given sample if, for any artefacts within that sample, the reconstruction can approximate (with large enough reconstruction error such that it is classified as an artefact) the `true' unknown underlying waveform behaviour that would be expected in the absence of the artefact-driving mechanism, whilst simultaneously maintaining a small error for any non-artefact regions within the same sample (e.g. Figure \ref{fig:latent_examples}a, left column and second row). The latent representation, which is described by the variational distribution $q_\phi(z|x)$, of an artefact gives some hint as to when DeepClean is successful at this. Inputs containing an artefact but with high probability mass in the latent distribution have `valid' synthetic waveforms that are indeed of the underlying process described by the generative model and we can therefore impute the section marked as an artefact with its reconstruction. Conversely, in imposing the generative model and a latent representation on new data that does not come from the same process or mechanism as the training data, in some cases the probability that an artefactual sample is generated by latent variables that are similar to those of the (valid) training data is very low and, as a result, the artefactual sample is forced to have a latent representation with vanishingly small probability mass in the average encoding distribution or the prior. As the model has spent little or no time training this region of the latent space, the reconstruction is therefore very unlike that of any valid data, and we cannot then impute using it (e.g. Figure \ref{fig:latent_examples}a, right column and second row). Density-based anomaly detection methods, which identify anomalies by the sparsity of the region of space in which they occupy, may be useful to recognise artefacts for which this is the case. An alternative in this case is to track the trajectory of a patient in the latent space and sample at points close to this trajectory when an artefact has been identified, but further work is needed to understand the structure of the latent space. For example, $\beta$-VAEs \cite{Higgins2017-lv} can improve the ability of the network to disentangle meaningful features in the data via constrained optimisation.

We have considered expert manual mark-up for `ground truth' artefact assignment. Some artefacts are clearly identifiable to clinicians because their waveform profile is so extreme. For example, blood sampling or subsequent flushing of the arterial line, which may be very variable in profile but will clearly contain unphysiological waveform excursions. There are other waveforms that are not so clear-cut to categorically assign as artefact. Patient movement may introduce vibration, which in the extreme may render the waveform unusable but in less severe cases often simply decreases the signal-to-noise ratio; whether such signals are identified as artefacts is to some extent arbitrary. Changes in the resonant properties of the ABP transduction system due to blood clots or bubbles represent another difficulty. This may occur as a result of `over-damping' (mean pressure preserved) or attenuation \cite{Ercole2006-zf} (mean pressure not preserved): in either case the pulse amplitude is reduced and high-frequency features are lost. Since the presence of high frequency features varies with cardiac output, it is impossible to absolutely ascribe such regions as valid or otherwise. As a result of these considerations, a good gold standard is lacking and therefore imposes limitations on the ability to rigorously evaluate model performance.

There are a number of improvements that could be made to our framework. We focused little on optimising hyperparameter and architecture choices, using deep CNNs for both encoder and decoder that we deliberately kept small relative to the deep generative learning literature. However, we expect the gains from increasing the network size or complexity, for example, to be marginal in this application domain. Indeed, increasing the learning capacity of the decode may mean that the latent variables are ignored and do not encode information about the data. One hyperparameter we have investigated is the latent dimension. This is particularly interesting in view of unsupervised representation learning, provided we are able to learn disentangled interpretable features encoded in the latent space.

We investigated relaxing the assumptions that the conditional (generative) distribution is a factorised Gaussian with identical variances. Further discussion of this assumption is given in \cite{Shu2018-tj}. Fixing this variance vector such that the components are all equal to a value $\sigma^2$ is equivalent to introducing a $\beta$ hyperparameter as in \cite{Higgins2017-lv}, with $\beta^{-1} = \sigma^2$. Instead, we split the decoder into two modules with identical architecture (matching the decoder module in Figure \ref{fig:network}), with the output of one becoming the mean of the multivariate Gaussian and the other becoming the log-variance terms of a diagonal covariance matrix, not necessarily with identical elements. This covariance offers, in theory, a more natural route to determining artefacts within the sample: given that for each input we have learnt parameters for a generative Gaussian distribution conditional upon its latent representation, we can easily calculate confidence regions or ellipsoids using, for example, the Mahalabonis distance \cite{Slotani1964-bx}. We can then determine the confidence level that the input belongs to its corresponding output Gaussian distribution under the assumed generative process, which is perhaps more intuitive than the method we used. For a binary classification, we can then reject samples according to some prescribed threshold. Within-sample artefact detection is possible by marginalising over a subset of the variables. However, though it makes sense to allow the network to quantify uncertainty in the reconstruction within each sample, it hinders the discriminative ability to identify artefacts, since the model assigns much higher variance to artefactual samples of a form that it has not previously seen in training. As a result, even though the distance between the input and reconstruction may be greater for artefactual samples, the confidence level of a ellipsoid containing the input might be higher than for normal waveforms. It is certainly possible to train the network in this way and still use the MSE or an equivalent metric to identify artefacts as before. In the case of diagonal covariance with identical elements, $\Sigma = \sigma^2 I$, the log-density is proportional to the MSE and the only difference between these approaches is the way in which we defined a threshold.

We chose the ABP waveform as a test case for several reasons. Firstly, it is particularly artefact-prone due to the effects of movement and flushing on the fluid filled catheter system typically employed. Secondly, ABP is of universal physiological importance: especially the extremes of ABP (hypo- and hypertension), which are particularly influenced by artefacts. Finally, ABP morphology tends to have good signal to noise properties. The performance of our model with other types of physiological waveforms is an area of future investigation and optimisation. We have trained and tested our system on available data, from a patient in sinus rhythm which is strongly periodic over the sample lengths considered. The performance of this framework on data from patients with more irregular (e.g. patients with arrhythmias, such as atrial fibrillation or multiple ectopics) has not been determined and will be the subject of future work. It is important to establish whether a generative model trained on the waveforms from one patient is able to generalise to other patients and maintain similar performance. Though training separate models for each patient is more cost-efficient than a manual annotation, it is obviously much more beneficial if artefact prediction is possible without requiring training data from each patient. If retraining is required, it is possible to initialise from a previous model and so not necessary to train from scratch. There may be practical complications in such transfer of learning between patients, such as ensuring the length of the samples corresponding to a fixed time interval is constant between patients, which would require either identical recording frequencies or an interpolation scheme.

This project offers a promising alternative to the identification of waveform artefacts in physiological waveforms from intensive care multimodality monitoring. We proposed a self-supervised generative deep learning approach to identify and reject these artefacts, DeepClean. This method demonstrated high sensitivity and specificity for identification of samples containing an artefact from an ICU ABP waveform. High-frequency waveforms are central to clinical prognosis of the patient state, and clinical parameters associated with outcome are derived from features of the signals. Analysis that may inform care and treatment is susceptible to biases that arise from unidentified signal artefacts, and subsequent potential misdiagnosis of clinical events could result in aggressive yet unnecessary clinical intervention. Further, by removing artefacts within the signal, any improvement in the prognostic ability of a real clinical event will reduce the high false positive rate of alarms in ICU monitoring systems, improving the conditions in which clinical staff can provide the best possible care for patients. Real-time identification of artefacts and signal irregularities is absolutely critical and this project suggests that generative deep learning can have an important role in this task.

\section*{Data and code availability}

The data used in this work were recorded as part of routine clinical care and not available for open access. However the authors will consider requests for collaborative research projects. The code used in this study is available at https://github.com/tedinburgh/deepclean. 

\section*{Author contributions}

T.E., S.E. and A.E. contributed to development of code and analysis. P.S. and M.C. contributed to the acquisition of data. T.E prepared the manuscript, which was edited by S.E. and A.E. and approved by all authors. Correspondence requests should be made to T.E. and A.E.

Competing interests: The authors declare no competing interests.

\section*{Acknowledgements}

Jim Stone for helpful discussion and Manuel Cabeleira for help with data preparation. T.E. is funded by Engineering and Physical Sciences Research Council (EPSRC) National Productivity Investment Fund (NPIF), reference 2089662, and Cantab Capital Institute for Mathematics of Information (CCIMI).

\bibliographystyle{ieeetr85}  
\bibliography{references1,references2}

\end{document}